\documentclass[runningheads]{llncs}
\usepackage{graphicx}
\usepackage{amsmath,amssymb} 
\usepackage[usenames,dvipsnames]{color}
\usepackage{empheq}
\usepackage{animate} 
\usepackage{array} 
\usepackage[english]{babel}
\usepackage[utf8]{inputenc}
\usepackage{algorithm}
\usepackage{multirow}
\usepackage[noend]{algpseudocode}
\usepackage{colortbl}%
\usepackage{gensymb}
\usepackage{diagbox}

\usepackage[width=122mm,left=12mm,paperwidth=146mm,height=193mm,top=12mm,paperheight=217mm]{geometry}
\usepackage[pagebackref=true,breaklinks=true,letterpaper=true,colorlinks,bookmarks=false]{hyperref}
\usepackage{arydshln}

\usepackage{xpatch}
\xapptocmd\normalsize{%
 \abovedisplayskip=10pt plus 3pt minus 9pt
 \abovedisplayshortskip=0pt plus 3pt
 \belowdisplayskip=10pt plus 3pt minus 9pt
 \belowdisplayshortskip=7pt plus 3pt minus 4pt
}{}{}

\begin{document}
\pagestyle{headings}
\mainmatter
\def\ACCV18SubNumber{429}  

\title{Maintaining Natural Image Statistics with the Contextual Loss
} 

\titlerunning{Maintaining Natural Image Statistics with the Contextual Loss}
\authorrunning{Roey Mechrez*, Itamar Talmi*, Firas Shama, Lihi Zelnik-Manor}
\newcommand*\samethanks[1][\value{footnote}]{\footnotemark[#1]}
\author{Roey Mechrez\thanks{indicate authors contributed equally}, Itamar Talmi\samethanks[1], Firas Shama, Lihi Zelnik-Manor}
\institute{The Technion - Israel\\
\email{\{roey@campus,titamar@campus,shfiras@campus,lihi@ee\}.technion.ac.il}}

\maketitle

\newcommand*{\ShowNotes}{}

\definecolor{darkred}{rgb}{0.7,0.1,0.1}
\definecolor{darkgreen}{rgb}{0.1,0.7,0.1}
\definecolor{cyan}{rgb}{0.7,0.0,0.7}
\definecolor{dblue}{rgb}{0.2,0.2,0.8}
\definecolor{maroon}{rgb}{0.76,.13,.28}
\definecolor{burntorange}{rgb}{0.81,.33,0}

\ifdefined\ShowNotes
  \newcommand{\colornote}[3]{{\color{#1}\bf{#2: #3}\normalfont}}
\else
  \newcommand{\colornote}[3]{}
\fi

\newcommand {\todo}[1]{\colornote{cyan}{TODO}{#1}}
\newcommand {\lihi}[1]{\colornote{magenta}{LZ}{#1}}
\newcommand {\itamar}[1]{\colornote{blue}{IT}{#1}}
\newcommand {\roey}[1]{\colornote{red}{RM}{#1}}
\newcommand {\firas}[1]{\colornote{cyan}{FS}{#1}}

\newcommand{\ignorethis } [1] {}
\newcommand{\DB         }     {{\mathcal{D}}}
\newcommand{\THR        }     {{\tau}}
\newcommand{\shortcite  }     {{\cite}}

\begin{figure}
\centering
\includegraphics[width=.92\linewidth]{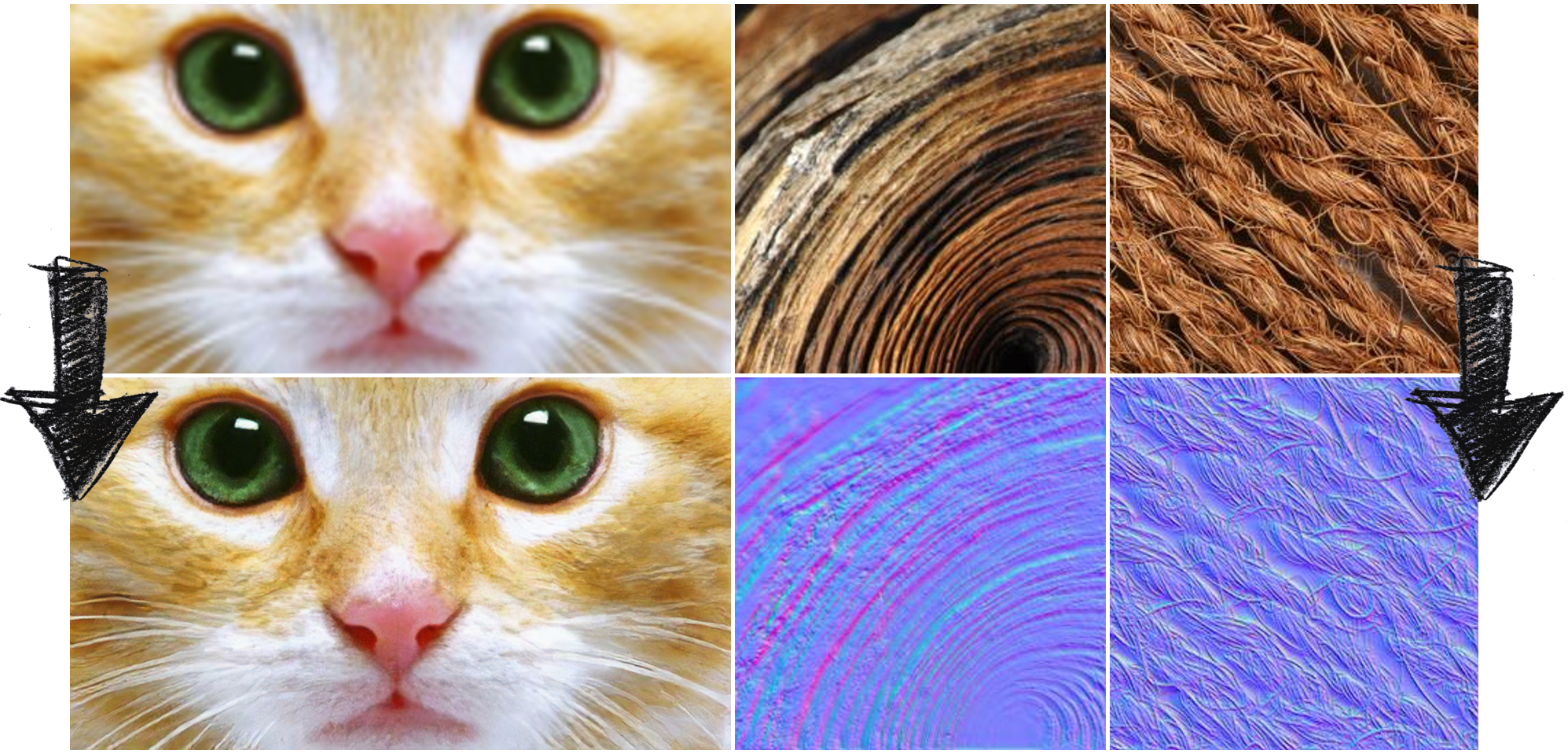}
\caption{We demonstrate the advantages of using a statistical loss for training a CNN in: (left) single-image super resolution, and, (right) surface normal estimation. Our approach is easy to train and yields networks that generate images (or surfaces) that exhibit natural internal statistics.}
\label{fig:teaser}
\end{figure}

\begin{abstract}
Maintaining natural image statistics is a crucial factor in restoration and generation of realistic looking images. When training CNNs, photorealism is usually attempted by adversarial training (GAN), that pushes the output images to lie on the manifold of natural images. GANs are very powerful, but not perfect. They are hard to train and the results still often suffer from artifacts. In this paper we propose a complementary approach, that could be applied with or without GAN, whose goal is to train a feed-forward CNN to maintain natural internal statistics. We look explicitly at the distribution of features in an image and train the network to generate images with natural feature distributions. Our approach reduces by orders of magnitude the number of images required for training and achieves state-of-the-art results on both single-image super-resolution, and high-resolution surface normal estimation.
\end{abstract}

\section{Introduction}
\begin{quote}
\emph{``Facts are stubborn things, but statistics are pliable.''}
― Mark Twain
\end{quote}


\noindent Maintaining natural image statistics has been known for years as a key factor in the generation of natural looking images~\cite{ruderman1994statistics,levin2003learning,weiss2007makes,roth2009fields,zoran2011learning}.
With the rise of CNNs, the utilization of explicit image priors was replaced by Generative Adversarial Networks (GAN)~\cite{goodfellow2014generative}, where a corpus of images is used to train a network to generate images with natural characteristics, e.g.,~\cite{radford2015unsupervised,arjovsky2017wasserstein,ledig2016photo,isola2016image}.
Despite the use of GANs within many different pipelines, results still sometimes suffer from artifacts, balanced training is difficult to achieve and depends on the architecture~\cite{arjovsky2017wasserstein,salimans2016improved}.



Well before the CNN era, natural image statistics were obtained by utilizing priors on the likelihood of the \emph{patches} of the generated images~\cite{levin2003learning,zoran2011learning,levin2007blind}.
Zoran and Weiss~\cite{zoran2011learning} showed that such statistical approaches, that harness priors on patches, lead in general to restoration of more natural looking images. 
A similar concept is in the heart of sparse-coding where a dictionary of visual code-words is used to constrain the generated patches~\cite{aharon2006rm,mairal2010online}.
The dictionary can be thought as a prior on the space of plausible image patches. 
A related approach is to constrain the generated image patches to the space of patches specific to the image to be restored~\cite{hassner2006example,michaeli2014blind,glasner2009super,zontak2011internal}.
In this paper we want to build on these ideas in order to answer the following question: \emph{Can we train a CNN to generate images that exhibit natural statistics?}




The approach we propose is a simple modification to the common practice.
As typically done, we as well train CNNs on pairs of source-target images by minimizing an objective that measures the similarity between the generated image and the target.
We extend on the common practice by proposing to use an objective that compares the \emph{feature distributions} rather than just comparing the appearance.
We show that by doing this the network learns to generate more natural looking images.

A key question is what makes a suitable objective for comparing distributions of features. 
The common divergence measures, such as Kullback-Leibler (KL), Earth-Movers-Distance and $\chi^2$, all require estimating the distribution of features, which is typically done via Multivariate Kernel-Density-Estimation (MKDE).
Since in our case the features are very high dimensional, and since we want a differentiable loss that can be computed efficiently, MKDE is not an option.
Hence, we propose instead an approximation to KL that is both simple to compute and tractable.
As it turns out, our approximation coincides with the recently proposed \textit{Contextual loss}~\cite{mechrez2018contextual}, that was designed for comparing images that are not spatially aligned.
Since the Contextual is actually an approximation to KL, it could be useful as an objective also in applications where the images are aligned, and the goal is to generate natural looking images.


Since all we propose is utilizing the Contextual loss during training, our approach is generic and can be adopted within many architectures and pipelines.
In particular, it can be used in concert with GAN training.
Hence, we chose super-resolution as a first test-case, where methods based on GANs are the current state-of-the-art.
We show empirically, that using our statistical approach with GAN training outperforms previous methods, yielding images that are \emph{perceptually} more realistic, while reducing the number of required training images by orders of magnitude.

Our second test-case further proves the generality of this approach to data that is not images, and shows the strength of the statistical approach without GAN. 
Maintaining natural internal statistics has been shown to be an important property also in the estimation of 3D surfaces~\cite{hassner2006example,gal2007surface,huang2000statistics}.
Hence, we present experiments on surface normal estimation, where the network's input is a high-resolution image and its output is a map of normals.
We successfully generate normal-maps that are more accurate than previous methods.

\noindent To summarize, the contributions we present are three-fold:
\begin{enumerate}
\item We show that the Contextual loss~\cite{mechrez2018contextual} can be viewed as an approximation to KL divergence. This makes it suitable for reconstruction problems.

\item We show that training with a statistical loss yields networks that maintain the natural internal statistics of images. This approach is easy to train and reduces significantly the training set size.

\item We present state-of-the-art  results on both perceptual super-resolution and high-resolution surface normal estimation.
\end{enumerate}


\section{Training CNNs to Match Image Distributions}
\label{sec:method}

\begin{figure}[t]
     	\centering
         \includegraphics[width=\linewidth]{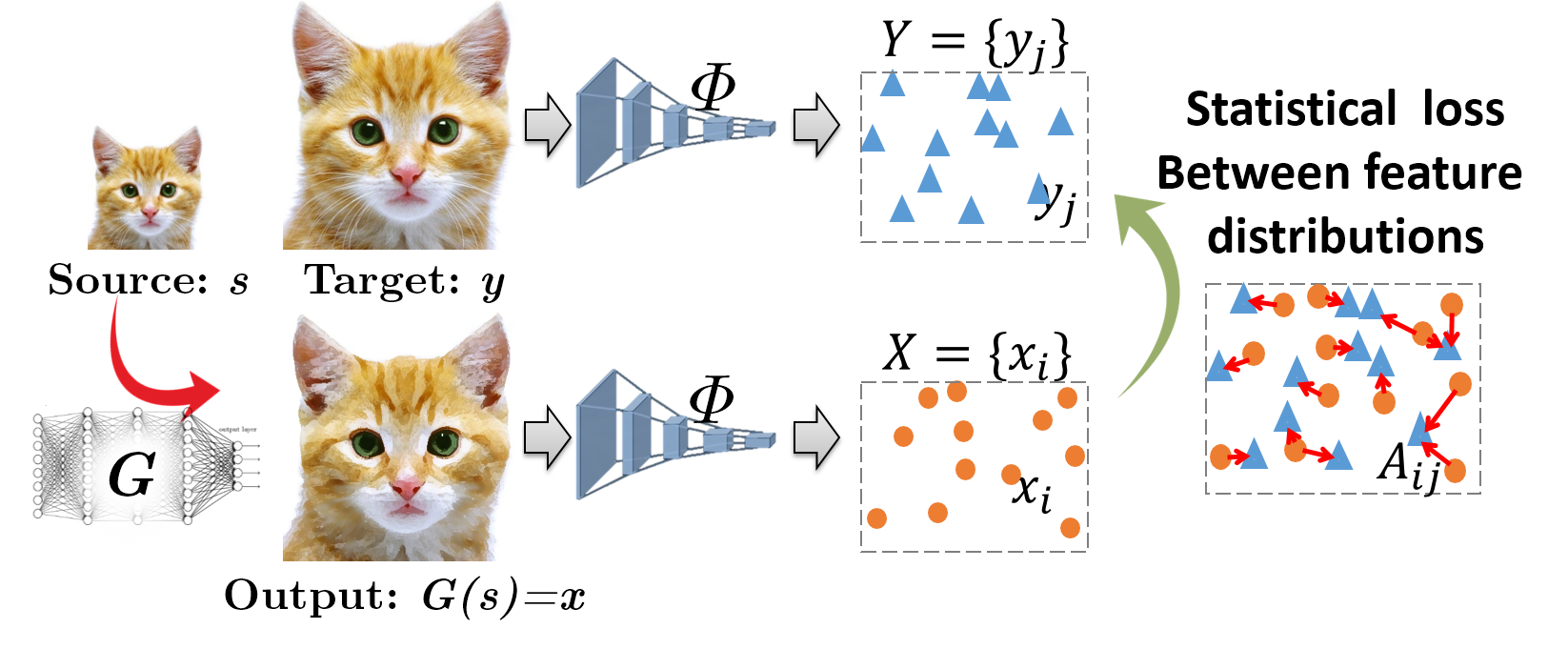}
        \vspace*{-0.4cm}
		\caption{\textbf{Training with a statistical loss}: 
        To train a generator network $G$ we compare the output image $G(s)\!\equiv\!x$ with the corresponding target image $y$ via a statistical loss --- the Contextual loss~\cite{mechrez2018contextual} --- that compares their feature distributions. 
        }
		\label{fig:CX}
\end{figure}

\subsection{Setup}
Our approach is very simple, as depicted in Figure~\ref{fig:CX}.
To train a generator network $G$ we use pairs of source $s$ and target $y$ images. The network outputs an image $G(s)\!\equiv\!x$, that should be as similar as possible to the target $y$.
To measure the similarity we extract from both $x$ and $y$ dense features $X\!=\!\{x_i\}$ and $Y\!=\!\{y_j\}$, respectively, e.g., by using pretrained network or vectorized RGB patches. 
We denote by $P_Y$ and $P_X$ the probability distribution functions over $Y$ and $X$,  respectively. When the patch set $X$ correctly models $Y$ then the underlying probabilities $P_Y$ and $P_X$ are equal. Hence, we need to train $G$ by minimizing an objective that measures the divergence between $P_X$ and $P_Y$.

\subsection{Review of the KL-divergence}
There are many common measures for the divergence between two distributions, e.g., $\chi^2$, Kullback-Leibler (KL), and Earth Movers Distance (EMD). 
We opted for the KL-divergence in this paper.

The KL-divergence between the two densities $P_X$ and $P_Y$ is defined as:
\newcommand\myeq{\stackrel{\mathclap{\normalfont\mbox{expectancy}}}{=}}
\begin{align}
KL(P_X||P_Y) = \int P_X \log \frac{P_X}{P_Y}
\end{align}
It computes the expectation of the logarithmic difference between the probabilities $P_X$ and $P_Y$, where the expectation is taken using the probabilities $P_X$.
We can rewrite this formula in terms of expectation:
\begin{align}
\label{eq:ExpectedKL}
KL(P_X||P_Y) = E\big[\log{P_X}-\log{P_Y}\big]
\end{align}
This requires approximating the parametrized densities ${P_X},{P_Y}$ from the feature sets $X\!=\!\{x_i\}$ and $Y\!=\!\{y_j\}$.
The most common method for doing this is to use Multivariate Kernel Density Estimation (MKDE), that estimates the probabilities $P_{X}(p_k)$ and $P_Y(p_k)$ as: 
\begin{align}
\label{eq:KDE}
P_{X}(p_k) = \sum_{x_i\in X} K_H(p_k,x_i) & \ \ \ \ \ 
P_{Y}(p_k) = \sum_{y_j\in Y} K_H(p_k,y_j) 
\end{align}
Here $K_H(p_k,\xi)$ is an affinity measure (the kernel) between some point $p_k$ and the samples $\xi$, typically taken as standard multivariate normal kernel $K_H(z)=(2\pi)^{-d/2}|H|^{-1/2}\exp(-\frac{1}{2}z^THz)$, with $z=dist(p_k,\xi)$, $d$ is the dimension and $H$ is a $d\!\times\!d$ bandwidth matrix.
We can now write the expectation of the KL-Divergence over the MKDE with uniform sample grid as follows: 
\begin{align}
\label{eq:KLsum}
KL(P_X||P_Y) = \frac{1}{N}\sum_{p_k} \big[\log{P_X(p_k)}-\log{P_Y(p_k)}\big] 
\end{align}

A common simplification that we adopt is to compute the MKDE not over a regular grid of $\{p_k\}$ but rather on the samples $\{x_i\}$ directly, i.e., we set $p_k\!=\!x_k$, which implies $K_H(p_k,x_i)\!=\!K_H(x_k,x_i)$ and $K_H(p_k,y_j)\!=\!K_H(x_k,y_j)$. 
Putting this together with Eq.~\eqref{eq:KDE} into Eq.~\eqref{eq:KLsum} yields:
\begin{align}
KL(P_X||P_Y) = \frac{1}{N}\sum_k 
\big[\log{\sum_{x_i\in X} K_H(x_k,x_i)} -\log{\sum_{y_j\in Y} K_H(x_k,y_j)}\big]
\label{eq:KDEinKL}
\end{align}

To use  Eq.~\eqref{eq:KDEinKL} as a loss function we need to choose a kernel $K_H$. 
The most common choice of a standard multivariate normal kernel requires setting the bandwidth matrix $H$, which is non-trivial. Multivariate KDE is known to be sensitive to the optimal selection of the bandwidth matrix and the existing solutions for tuning it require solving an optimization problem which is not possible as part of a training process. 
Hence, we next propose an approximation, which is both practical and tractable.


\subsection{Approximating the KL-divergence with the Contextual loss}

Our approximation is based on one further observation. It is insufficient to produce an image with the same distribution of features, but rather we want additionally that the samples will be similar. That is, we ask each point $x_i\in X$ to be close to a specific $y_j \in Y$. 

To achieve this, while assuming that the number of samples is large, we set the MKDE kernel such that it approximates a delta function. 
When the kernel is a delta, the first log term of Eq.~\eqref{eq:KDEinKL} becomes a constant since:
\begin{align}
K_H(x_k,x_i) = 
\begin{cases}
  \approx 1 & \text{if\ \ } k=i\\    
  \approx 0 &  \text{otherwise}
\end{cases}
\end{align}
The kernel in the second log term of Eq.~\eqref{eq:KDEinKL} becomes:
\begin{align}
K_H(x_k,y_j) =
\begin{cases}
  \approx 1  & \text{if\ \ } dist(x_k,y_j) \ll dist(x_k,y_l) \ \forall l\neq j\\   \approx 0 &  \text{otherwise}
\end{cases}
\label{eq:approxK}
\end{align}
We can thus simplify the objective of Eq.~\eqref{eq:KDEinKL}:
\begin{align}
\label{eq:obj}
E(x,y) = -\log \bigg( \, \frac{1}{N}\sum_{j} { \max_{i}{{A}_{ij}} }   \, \bigg)
\end{align} 
where we denote ${A}_{ij}=K_H(x_k,y_j)$.
Next, we suggest two alternatives for the kernel ${A}_{ij}$, and show that one implies that the objective of Eq.~\eqref{eq:obj} is equivalent to the Chamfer Distance~\cite{barrow1977parametric}, while the other implies it is equivalent to the Contextual loss of~\cite{mechrez2018contextual}.

\subsubsection{The Contextual Loss:}
As it turns out, the objective of Eq.~\eqref{eq:obj} is identical to the Contextual loss recently proposed by~\cite{mechrez2018contextual}.
Furthermore, they set the kernel ${A}_{ij}$ to be close to a delta function, such that it fits  Eq.~\eqref{eq:approxK}.
First the Cosine (or L2) distances $dist(x_i,y_j)$ are computed between all pairs $x_i$,$y_j$.
The distances are then normalized: $\tilde{d}_{ij}\!=\!{dist(x_i,y_j)}/({\min_{k}{dist(x_i,y_k)} + \epsilon})$ (with $\epsilon\!=\!1\mathrm{e}{-5}$), and finally the pairwise affinities $A_{ij}\!\in\![0,1]$ are defined as: 
\vspace*{-0.1cm}
\begin{align}
\label{eq:Aij}
A_{ij} = \frac{\exp{\left( {1 - \tilde{d}_{ij}}/{h} \right)}}{\sum_{l}{\exp{\left( {1 - \tilde{d}_{il}}/{h} \right)}}}
= 
\begin{cases}
  \approx 1 & \text{if\ \ } \tilde{d}_{ij}\!\ll\!\tilde{d}_{il} \mbox{\ \ } \forall l\neq j\\    
  \approx 0 &  \text{otherwise}
\end{cases}
\end{align}
where $h\!>\!0$ is a scalar bandwidth parameter that we fix to $h\!=\!0.1$, as proposed in~\cite{mechrez2018contextual}. 
When using these affinities our objective equals the Contextual loss of~\cite{mechrez2018contextual} and we denote $E(x,y) = \mathcal{L}_{\text{CX}}(x,y)$.


\subsubsection{The Chamfer Distance}
A simpler way to set ${A}_{ij}$ is to take a Gaussian kernel with a fixed $H=I$ s.t. $A_{ij}=exp(-dist(x_i,y_j))$. 
This choice implies that minimizing Eq.~\eqref{eq:obj} is equivalent to  minimizing the asymmetric Chamfer Distance~\cite{barrow1977parametric} between $X,Y$ defined as
\begin{align}
\text{CD}(X,Y) = \frac{1}{|X|}\sum_{i} \min_{j} dist(x_i,y_j) 
\end{align}
CD has been previously used mainly for shape retrieval~\cite{sun2018pix3d,de2012vision} were the points are in $\mathbb{R}^3$. 
For each point in set $X$, CD finds the nearest point in the set $Y$ and minimizes the sum of these distances. 
A downside of this choice for the kernel ${A}_{ij}$ is that it does not satisfy the requirement in Eq.~\eqref{eq:approxK}.

\begin{figure}[t]
		\setlength{\tabcolsep}{.8em}
     	\centering
        \begin{tabular}{ccc}
        \includegraphics[width=0.25\linewidth]{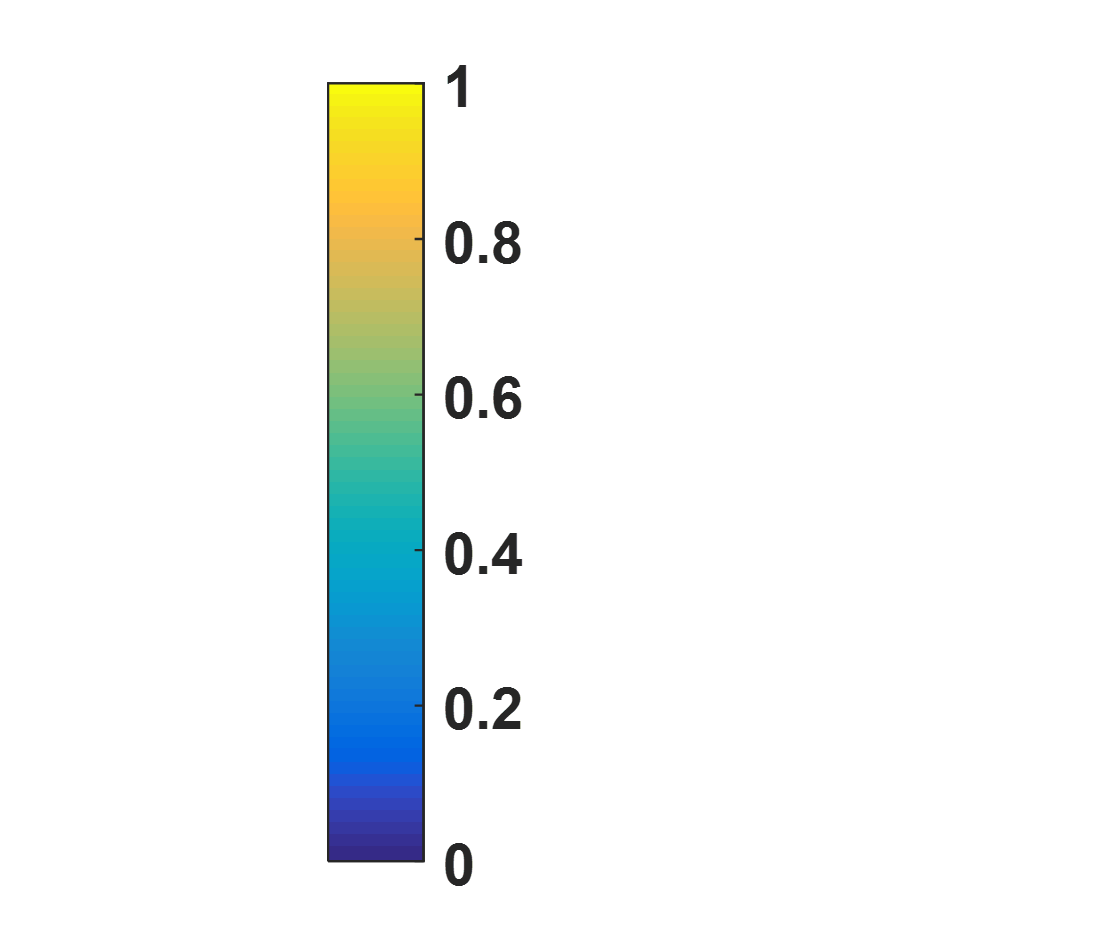}&
        \includegraphics[width=0.3\linewidth]{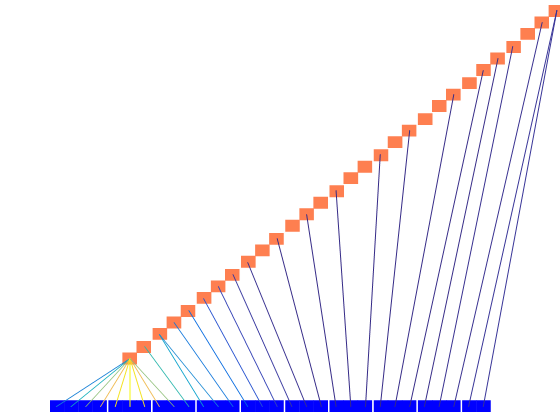}&
        \includegraphics[width=0.3\linewidth]{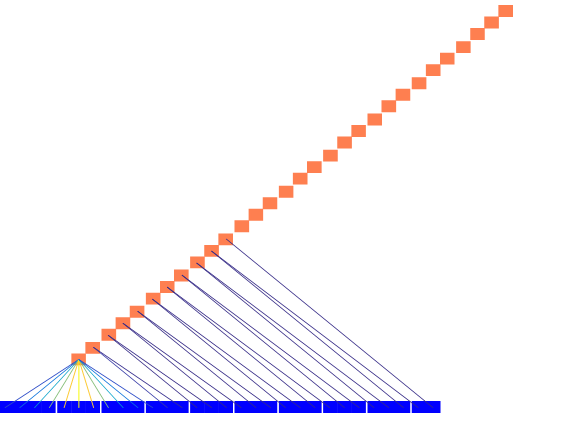}\\
        Affinity value &
        (a) The Contextual loss~\cite{mechrez2018contextual} & (b) Chamfer distance~\cite{barrow1977parametric}
        \end{tabular}
         
		\caption{\textbf{The Contextual loss vs. Chamfer Distance}: We demonstrate via a 2D example the difference between two approximations to the KL-divergence (a) The Contextual loss and (b) Chamfer Distance. Point sets $Y$ and $X$ are marked with blue and orange squares respectively. The colored lines connect each $y_j$ with $x_i$ with the largest affinity.
        The KL approximation of Eq.~\eqref{eq:obj} sums over these affinities.
        It can be seen that the normalized affinities used by the Contextual loss lead to more diverse and meaningful matches between $Y$ and $X$.
        }
		\label{fig:dots}
\end{figure}
\subsubsection{The Contextual Loss vs. Chamfer Distance}
To provide intuition on the difference between the two choices of affinity functions we present in Figure~\ref{fig:dots} an illustration in 2D.
Recall, that both the Contextual loss $\mathcal{L}_{\text{CX}}$ and the Chamfer distance CD find for each point in $Y$ a single match in $X$, however, these matches are computed differently.
CD selects the closet point, hence, we could get that multiple points in $Y$ are matched to the same few points in $X$.
Differently, $\mathcal{L}_{\text{CX}}$ computes normalized affinities, that consider the distances between every point $x_i$ to all the points $y_j\!\in\!Y$. 
Therefore, it results in more diverse matches between the two sets of points, and provides a better measure of similarity between the two distributions. Additional example is presented in the supplementary.

Training with $\mathcal{L}_{\text{CX}}$ guides the network to match between the two point sets, and as a result the underlying distributions become closer. In contrast, training with CD, does not allow the two sets to get close. Indeed, we found empirically that training with CD does not converge, hence, we excluded it from our empirical reports.


\section{Empirical analysis}

The Contextual loss has been proposed in~\cite{mechrez2018contextual} for measuring similarity between non-aligned images. It has therefore been used for applications such as style transfer, where the generated image and the target style image are very different.
In the current study we assert that the Contextual loss can be viewed as a statistical loss between the distributions of features. 
We further assert that using such a loss during training would lead to generating images with realistic characteristics. 
This would make it a good candidate also for tasks where the training image pairs are aligned.

To support these claims we present in this section two experiments, and in the next section two real applications.
The first experiment shows that minimizing the Contextual loss during training indeed implies also minimization of the KL-divergence.
The second experiment evaluates the relation between the Contextual loss and human perception of image quality.

\subsubsection{Empirical analysis of the approximation}
\begin{figure}[t]
\centering
\includegraphics[width=\textwidth]{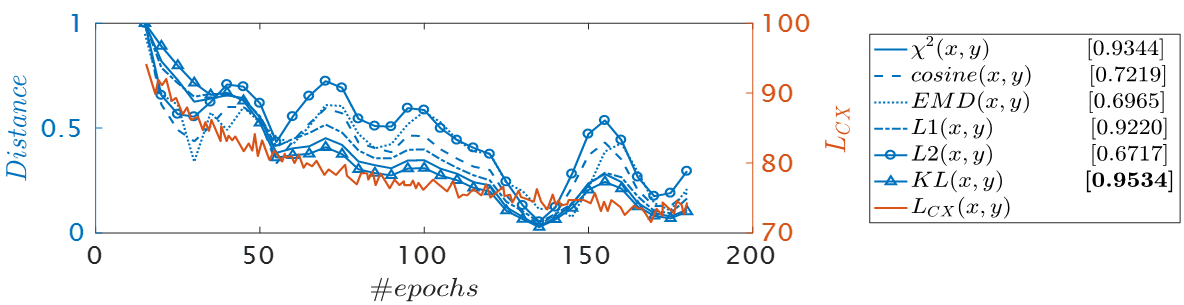}
\vspace*{-0.2cm}
\caption{\textbf{Minimizing the KL-divergence:} 
The curves represents seven measures of dissimilarity between the patch distribution of the target image $y$ and that of the generated image $x$, during training a super-resolution network (details in supplementary) and using $\mathcal{L}_{\text{CX}}$ as the objective.
The correlation between $\mathcal{L}_{\text{CX}}$ and all other measures, reported in square brackets, show highest correlation to the KL-divergence.
It is evident that training with the Contextual loss results in minimizing also the KL-divergence.}
\label{fig:kde}
\end{figure}

Since we are proposing to use the Contextual loss as an approximation to the KL-divergence, we next show empirically, that minimizing it during training also minimizes the KL-divergence.

To do this we chose a simplified super-resolution setup, based on SRResNet~\cite{ledig2016photo}, and trained it with the Contextual loss as the objective. 
The details on the setup are provided in the supplementary.
We compute during the training the Contextual loss, the KL-divergence, as well as five other common dissimilarity measures.
To compute the KL-divergence, EMD and $\chi^2$, we need to approximate the density of each image.
As discussed in Section~\ref{sec:method}, it is not clear how the multivariate solution to KDE can be smoothly used here, therefore, instead, we generate a random projection of all $5\!\times\!5$ patches onto 2D and fit them using KDE with a Gaussian kernel in 2D~\cite{botev2010kernel}. 
This was repeated for 100 random projections and the scores were averaged over all projections (examples of projections are shown in the supplementary).

Figure~\ref{fig:kde} presents the values of all seven measures during the iterations of the training.
It can be seen that all of them are minimized during the iterations.
The KL-divergence minimization curve is the one most similar to that of the Contextual loss, suggesting that the Contextual loss forms a reasonable approximation.

\subsubsection{Evaluating on Human Perceptual Judgment}
\begin{figure}[t]
     	\centering
         \includegraphics[width=\linewidth]{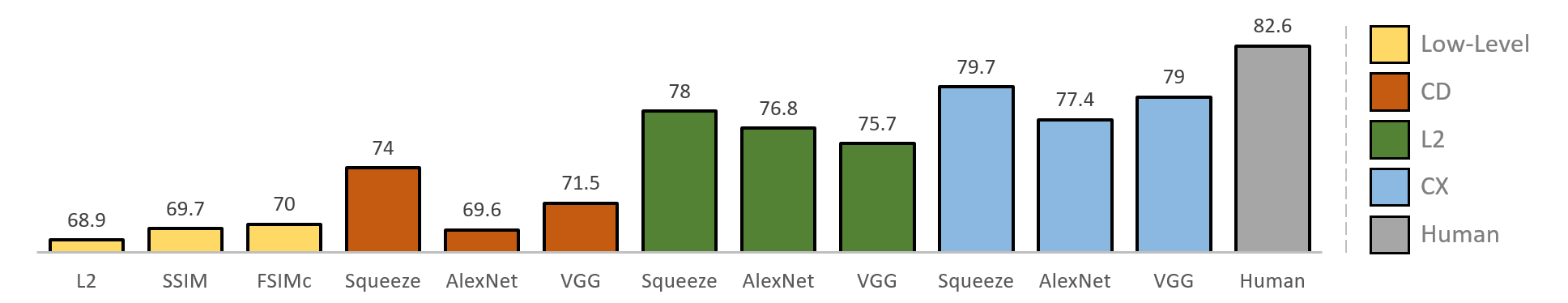}
        \vspace*{-0.4cm}
		\caption{\textbf{How perceptual is the Contextual loss?}
        Test on the 2AFC~\cite{zhang2018unreasonable} data set with traditional and CNN-based distortions. We compare between low-level metrics (such as SSIM) and off-the-shelf deep networks trained with L2 distance, Chamfer distance (CD) and the Contextual loss ($\mathcal{L}_{\text{CX}}$).  $\mathcal{L}_{\text{CX}}$ yields a performance boost across all three networks with the largest gain obtained for VGG -- the most common loss network (i.e., perceptual loss). }
		\label{fig:perceptual}
\end{figure}

Our ultimate goal is to train networks that generate images with high perceptual quality. The underlying hypothesis behind our approach is that training with an objective that maintains natural statistics will lead to this goal. 
To asses this hypothesis we repeated the evaluation procedure proposed in~\cite{zhang2018unreasonable}, for assessing the correlation between human judgment of similarity and loss functions based on deep features.

In~\cite{zhang2018unreasonable} it was suggested to compute the similarity between two images by comparing their corresponding deep embeddings.
For each image they obtained a deep embedding via a pre-trained network, normalized the activations and then computed the $L2$ distance.
This was then averaged across spatial dimension and across all layers. 
We adopted the same procedure while replacing the $L2$ distance with the Contextual loss approximation to KL.

Our findings, as reported in Figure~\ref{fig:perceptual} (the complete table is provided in the supplementary material), show the benefits of our proposed approach.
The Contextual loss between deep features is more closely correlated with human judgment than $L2$ or Chamfer Distance between the same features.
All of these perceptual measures are preferable over low-level similarity measures such as SSIM~\cite{wang2004image}.





\section{Applications}

In this section we present two applications:  single-image super-resolution, and high-resolution surface normal estimation.
We chose the first to highlight the advantage of using our approach in concert with GAN.
The second was selected since its output is not an image, but rather a surface of normals. This shows the generic nature of our approach to other domains apart from image generation, where GANs are not being used.


\subsection{Single-Image Super-Resolution}
\label{sec:super-resolution}

To asses the contribution of our suggested framework for image restoration we experiment on single-image super-resolution. 
To place our efforts in context we start by briefly reviewing some of the recent works on super-resolution. A more comprehensive overview of the current trends can be found in~\cite{timofte2017NTIRE}. 

Recent solutions based on CNNs can be categorized into two groups.
The first group relies on the $L2$ or $L1$ losses~\cite{ledig2016photo,Lim2017edsr,tong2017image}, which lead to high PSNR and SSIM~\cite{wang2004image} at the price of low perceptual quality, as was recently shown in~\cite{blau2017perception,zhang2018unreasonable}. 
The second group of works aim at high perceptual quality.
This is done by adopting perceptual loss functions~\cite{johnson2016perceptual}, sometimes in combination with GAN~\cite{ledig2016photo} or by adding the Gram loss~\cite{gatys2016image}, which nicely captures textures~\cite{sajjadi2017enhancenet}.

\paragraph{\textbf{Proposed solution:} \ \ \ } 
Our main goal is to generate natural looking images, with natural internal statistics.
At the same time, we do not want the structural similarity to be overly low (the trade-off between the two is nicely analyzed in~\cite{blau2017perception,zhang2018unreasonable}).   
Therefore, we propose an objective that considers both, with higher importance to perceptual quality. 
Specifically, we integrate three loss terms:
(i) The Contextual loss -- to make sure that the internal statistics of the generated image are similar to those of the ground-truth high-resolution image.
(ii) The $L2$ loss, computed at \textbf{low} resolution -- to drive the generated image to share the spatial structure of the target image. 
(iii) Finally, following~\cite{ledig2016photo} we add an adversarial term, which helps in pushing the generated image to look ``real''. 

Given a low-resolution image $s$ and a target high-resolution image $y$, our objective for training the network $G$ is:
\begin{equation}
\mathcal{L}(G) = \lambda_{\text{CX}}\cdot\mathcal{L}_{\text{CX}}(G(s),y) + \lambda_{{L2}}\cdot ||G(s)^{\text{LF}}-y^{\text{LF}}||_2 +\lambda_{\text{GAN}}\cdot\mathcal{L}_{\text{GAN}}(G(s))
\end{equation}
where in all our experiments $\lambda_{\text{CX}}\!=\!0.1$, $\lambda_{\text{GAN}}\!=\!1e-3$, and $\lambda_{{L2}}\!=\!10$.
The images $G(s)^{\text{LF}},y^{\text{LF}}$ are low-frequencies obtained by convolution with a Gaussian kernel of width $21\!\times\!21$ and $\sigma\!=\!3$. For the Contextual loss feature extraction we used layer $conv3\_4$ of VGG19~\cite{simonyan2014very}.

\paragraph{\textbf{Implementation details:} \ \ \ } 
We adopt the SRGAN architecture~\cite{ledig2016photo}\footnote{We used the implementation in \url{https://github.com/tensorlayer/SRGAN}} and replace only the objective.
We train it on just 800 images from the DIV2K dataset~\cite{agustsson2017div2k}, for 1500 epochs. Our network is initialized by first training using only the $L2$ loss for 100 epochs. 

\paragraph{\textbf{Evaluation:}\ \ \  } 
Empirical evaluation was performed on the BSD100 dataset~\cite{martin2001database}.
As suggested in~\cite{blau2017perception} we compute both structural similarity (SSIM~\cite{wang2004image}) to the ground-truth and perceptual quality (NRQM~\cite{ma2017learning}).
The ``ideal'' algorithm will have both scores high.

Table~\ref{fig:SR_table} compares our method with three recent solutions whose goal is high perceptual quality. 
It can be seen that our approach outperforms the state-of-the-art on both evaluation measures.
This is especially satisfactory as we needed only $800$ images for training, while previous methods had to train on tens or even hundreds of thousands of images.
Note, that the values of the perceptual measure NRQM are not normalized and small changes are actually quite significant. For example the gap between \cite{ledig2016photo} and \cite{sajjadi2017enhancenet} is only 0.014 yet visual inspection shows a significant difference. The gap between our results and \cite{sajjadi2017enhancenet} is 0.08, i.e., almost 6 times bigger, and visually it is significant.
\begin{table}[t]
\setlength{\tabcolsep}{.7em}
\centering
\scalebox{0.99}{%
\begin{tabular}{l | ccc : c}
\multirow{2}{*}{\textbf{Method}}  & \textbf{Loss} & \textbf{Distortion} & \textbf{Perceptual} &  \# Training\\
& \textbf{function} & \textbf{SSIM}~\cite{wang2004image} & \textbf{NRQM}~\cite{ma2017learning} & images\\
\hline
\hline
Johnson\cite{johnson2016perceptual} & $L_P$ & 0.631 & 7.800 & 10K \\
SRGAN\cite{ledig2016photo} & $\mathcal{L}_{\text{GAN}}$+$L_P$ & 0.640 & 8.705 & 300K \\
EnhanceNet\cite{sajjadi2017enhancenet} & $\mathcal{L}_{\text{GAN}}$+$L_P$+$L_T$ & 0.624 & 8.719 & 200K \\
Ours \textbf{full} & $\mathcal{L}_{\text{GAN}}$+$\mathcal{L}_{\text{CX}}$+$L_2^{LF}$ & {0.643} & \textbf{8.800} & 800\\
\hline
 Ours w/o $L_2^{LF}$ & $\mathcal{L}_{\text{GAN}}$+$\mathcal{L}_{\text{CX}}$ &\textbf{0.67} & 8.53 & 800\\
Ours w/o $\mathcal{L}_{\text{CX}}$ & $\mathcal{L}_{\text{GAN}}$+$L_2^{LF}$ & 0.510 & 8.411 & 800\\
SRGAN-MSE* & $\mathcal{L}_{\text{GAN}}$+$L_2$ & 0.643 & 8.4 & 800 \\
\multicolumn{4}{l}{\small *our reimplementation} \\
\end{tabular}
}
\vspace*{0.1cm}
\caption{\textbf{Super-resolution results}: 
The table presents the mean scores obtained by our method and three others on BSD100~\cite{martin2001database}. 
SSIM~\cite{wang2004image} measures structural similarity to the ground-truth, while NRQM~\cite{ma2017learning} measures no-reference perceptual quality.
Our approach provides an improvement on both scores, even though it required orders of magnitude fewer images for training. The table further provides an ablation test of the loss function to show that $\mathcal{L}_{\text{CX}}$ is a key ingredient in the quality of the results. $L_P$ is the perceptual loss~\cite{johnson2016perceptual}, $L_T$ is the Gram loss~\cite{gatys2016image}, and both use VGG19 features. 
}
\vspace*{-0.5cm}
\label{fig:SR_table}
\end{table}

\begin{figure}[t]
	\centering
	\small
        \setlength{\tabcolsep}{.15em}
        \footnotesize
     \begin{tabular}{cccc|c} 
    \includegraphics[width=.19\linewidth]{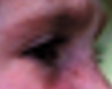}&
    \includegraphics[width=.19\linewidth]{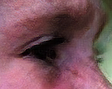}&
    \includegraphics[width=.19\linewidth]{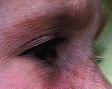}&
    \includegraphics[width=.19\linewidth]{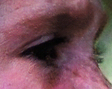}&
     \includegraphics[width=.19\linewidth]{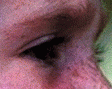}\\
    \includegraphics[width=.19\linewidth]{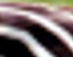}&
    \includegraphics[width=.19\linewidth]{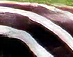}&
    \includegraphics[width=.19\linewidth]{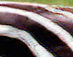}&
    \includegraphics[width=.19\linewidth]{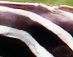}&
    \includegraphics[width=.19\linewidth]{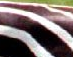}\\
    \includegraphics[width=.19\linewidth]{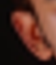}&
    \includegraphics[width=.19\linewidth]{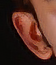}&
    \includegraphics[width=.19\linewidth]{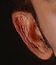}&
    \includegraphics[width=.19\linewidth]{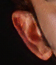}&
    \includegraphics[width=.19\linewidth]{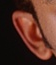}\\
     Bicubic  & EnhanceNet\cite{sajjadi2017enhancenet} & SRGAN\cite{ledig2016photo} & ours & HR
	\end{tabular}
    \vspace*{-0.19cm}
    \caption{
		\textbf{Super-resolution qualitative evaluation}: 
        Training with GAN and a perceptual loss sometimes adds undesired textures, 
        e.g., SRGAN added wrinkles to the girl's eyelid, 
        unnatural colors to the zebra stripes, 
        and arbitrary patterns to the ear. 
        Our solution replaces the perceptual loss with the Contextual loss, thus getting rid of  the unnatural patterns.}
        \vspace*{-0.5cm}
	\label{fig:SR_results}
\end{figure}

Figure~\ref{fig:SR_results} further presents a few qualitative results, that highlight the gap between our approach and previous ones.
Both SRGAN~\cite{ledig2016photo} and EnhanceNet~\cite{sajjadi2017enhancenet} rely on adversarial training (GAN) in order to achieve photo-realistic results.
This tends to over-generate high-frequency details, which make the image look sharp, however, often these high-frequency components do not match those of the target image. 
The Contextual loss, when used in concert with GAN, reduces these artifacts, and results in natural looking image patches.

An interesting observation is that we achieve high perceptual quality while using for the Contextual loss features from a mid-layer of VGG19, namely $conv3\_4$.
This is in contrast to the reports in~\cite{ledig2016photo} that had to use for SRGAN the high-layer $conv5\_4$ for the perceptual loss (and failed when using low-layer such as $conv2\_2$).
Similarly, EnhanceNet required a mixture of $pool2$ and $pool5$.

\subsection{High-resolution Surface Normal Estimation}
\label{sec:normals}

\begin{figure}[t]
		\centering
		\setlength{\tabcolsep}{.12em}
        \begin{tabular}{ccccc} 
		\includegraphics[width=.19\textwidth]{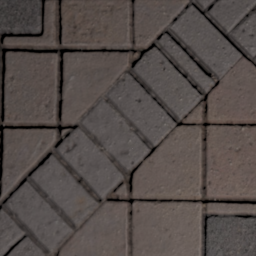}&
		\includegraphics[width=.19\textwidth]{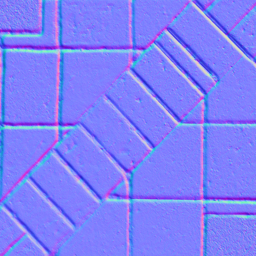}&
		\includegraphics[width=.19\textwidth]{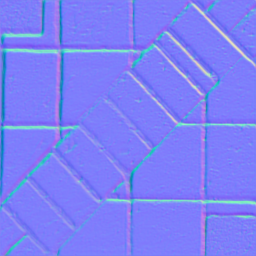}&
        \includegraphics[width=.19\textwidth]{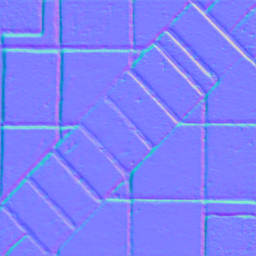}&
        \includegraphics[width=.19\textwidth]{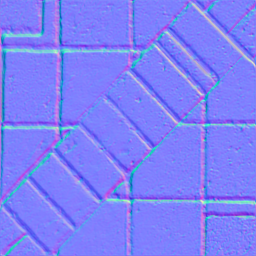}\\
        \includegraphics[width=.19\textwidth]{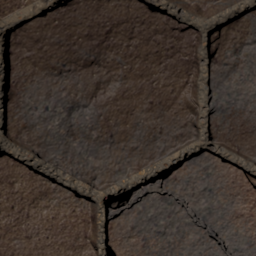}&
		\includegraphics[width=.19\textwidth]{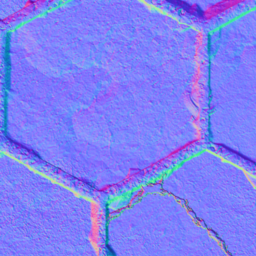}&
        \includegraphics[width=.19\textwidth]{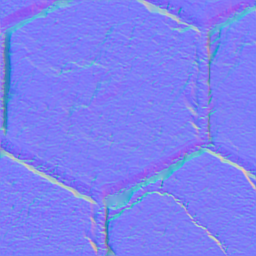}&
        \includegraphics[width=.19\textwidth]{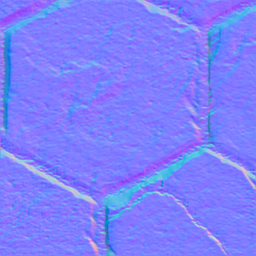}&
        \includegraphics[width=.19\textwidth]{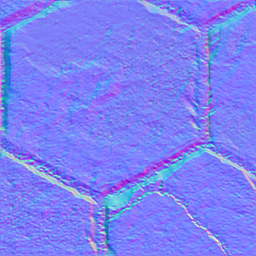}\\
        %
        Texture image & Ground-truth  & CRN & Ours (w/ $L1$) & Ours (w/o $L1$)\\
        \end{tabular}
		\caption{\textbf{Estimating surface normals}: A visual comparison of our method with CRN~\cite{chen2017photographic}. CRN trains with $L1$ as loss, which leads to coarse reconstruction that misses many fine details. Conversely, using our objective, especially without $L1$, leads to more natural looking surfaces, with delicate texture details. 
        Interestingly, the results without $L1$ look visually better, even though quantitatively they are inferior, as shown in Table~\ref{fig:normal_table}. We believe this is due to the usage of structural evaluation measures rather than perceptual ones.
        }
		\label{fig:normal_data}
\end{figure}

The framework we propose is by no means limited to networks that generate images. It is a generic approach that could be useful for training networks for other tasks, where the natural statistics of the target should be exhibited in the network's output.
To support this, we present a solution to the problem of surface normal estimation -- an essential problem in computer vision, widely used for scene understanding and new-view synthesis. 

\paragraph{\textbf{Data:}\ \ \ }
The task we pose is to estimate the underlying normal map from a single monocular color image. Although this problem is ill-posed, recent CNN based approaches achieve satisfactory results ~\cite{bansal2017pixelnet,wang2015designing,eigen2015predicting,chen2017surface} on the NYU-v2 dataset~\cite{silbermanECCV12NYU}. 
Thanks to its size, quality and variety, NYU-v2 is intensively used as a test-bed for predicting depth, normals, segmentation etc.
However, due to the acquisition protocol, its data does not include the fine details of the scene and misses the underlying high frequency information, hence, it does not provide a detailed representation of natural surfaces.

Therefore, we built a new dataset of images of surfaces and their corresponding normal maps, where fine details play a major role in defining the surface structure. 
Examples are shown in Figure~\ref{fig:normal_data}.
Our dataset is based on 182 different textures and their respective normal maps that were collected from the Internet\footnote{\href{www.poliigon.com}{www.poliigon.com} and \href{www.textures.com}{www.textures.com}}, originally offered for usages of realistic interior home design, gaming, arts, etc.
For each texture we obtained a high resolution color image ($1024\!\times\!1024$) and a corresponding normal-map of a surface such that its underlying plane normal points towards the camera (see Figure~\ref{fig:normal_data}). 
Such image-normals pairs lack the effect of lighting, which plays an essential role in normal estimation.
Hence, we used Blender\footnote{\href{www.blender.org}{www.blender.org}}, a 3D renderer, to simulate each texture under $10$ different point-light locations. This resulted in a total of $1820$ pairs of image-normals.
The textures were split into $90\%$ for training and $10\%$ for testing, such that the test set includes all the $10$ rendered pairs of each included texture.

The collection offers a variety of materials including wood, stone, fabric, steel, sand, etc., with multiple patterns, colors and roughness levels, that capture the appearance of real-life surfaces. 
While some textures are synthetic, they look realistic and exhibit imperfections of real surfaces, which are translated into the fine-details of both the color image as well as its normal map.

\paragraph{\textbf{Proposed solution:}\ \ \ }
We propose using an objective based on a combination of three loss terms:
(i) The Contextual loss -- to make sure that the internal statistics of the generated normal map match those of the target normal map.
(ii) The $L2$ loss, computed at \textbf{low} resolution, and (iii) The $L1$ loss. Both drive the generated normal map to share the spatial layout of the target map.  
Our overall objective is:
\begin{equation}
\label{eq:normal-loss}
\mathcal{L}(G) = \lambda_{\text{CX}}\cdot\mathcal{L}_{\text{CX}}(G(s),y) + \lambda_{{L2}}\cdot ||G(s)^{\text{LF}}-y^{\text{LF}}||_2 + \lambda_{L1}\cdot ||G(s)-y||_1
\end{equation}
where, $\lambda_{\text{CX}}\!=\!1$, and $\lambda_{{L2}}\!=\!0.1$. The normal-maps $G(s)^{\text{LF}},y^{\text{LF}}$ are low-frequencies obtained by convolution with a Gaussian kernel of width $21\!\times\!21$ and $\sigma=3$. 
We tested with both $\lambda_{L1}\!=\!1$ and $\lambda_{L1}\!=\!0$, which removes the third term.

\paragraph{\textbf{Implementation details:} \ \ \ } 
We chose as architecture the Cascaded Refinement Network (CRN)~\cite{chen2017photographic} originally suggested for label-to-image and was shown to yield great results in a variety of other tasks~\cite{mechrez2018contextual}. 
For the contextual loss we took as features $5\!\times\!5$ patches of the normal map (extracted with stride 2) and layers $conv1\_2, conv2\_2$ of VGG19. In our implementation we reduced memory consumption by random sampling of all three layers into $65\!\times\!65$ features.

\paragraph{\textbf{Evaluation:} \ \ \ }
Table~\ref{fig:normal_table} compares our results with previous solutions.
We compare to the recently proposed PixelNet~\cite{bansal2017pixelnet}, that presented state-of-the-art results on NYU-v2. 
Since PixelNet was trained on NYU-v2, which lacks fine details, we also tried fine-tuning it on our dataset. 
In addition, we present results obtained with CRN~\cite{chen2017photographic}.
While the original CRN was trained with both $L1$ and the perceptual loss~\cite{johnson2016perceptual}, this combination provided poor results on normal estimation.
Hence, we excluded the perceptual loss and report results with only $L1$ as the loss, or when using our objective of Eq.~\eqref{eq:normal-loss}.
It can be seen that CRN trained with our objective (with $L1$) leads to the best quantitative scores.
\vspace*{-0.3cm}
\begin{table}[h]
\setlength{\tabcolsep}{.6em}
\centering
\begin{tabular}{l | cccccc}
 & Mean &  Median & RMSE & $11.25\degree$ & $22.5\degree$ & $30\degree$\\
\textbf{Method} & ($\degree$)$\downarrow$ &  ($\degree$)$\downarrow$ & ($\degree$)$\downarrow$ & (\%)$\uparrow$ & (\%)$\uparrow$ & (\%)$\uparrow$\\
\hline
PixelNet~\cite{bansal2017pixelnet} & 25.96&	23.76&	30.01&	22.54&	50.61&	65.09\\
PixelNet~\cite{bansal2017pixelnet}+FineTune & 14.27	&12.44&	16.64&	51.20&	85.13&	91.81\\
CRN~\cite{chen2017photographic} & 8.73&	6.57&	11.74&	74.03&	90.96&	95.28\\
Ours (without $L1$) & 9.67&	7.26&	12.94&	70.39&	89.84&	94.73\\
Ours (with $L1$) & \textbf{8.59}	&\textbf{6.50}&	\textbf{11.54}&	\textbf{74.61}&	\textbf{91.12}	&\textbf{95.28}\\
\end{tabular}

\vspace*{0.1cm}
\caption{\textbf{Quantitative evaluation of surface normal estimation} (on our new high-resolution dataset).
The table presents six evaluation statistics, previously suggested by~\cite{bansal2017pixelnet,wang2015designing}, over the angular error between the predicted normals and ground-truth normals.
For the first three criteria lower is better, while for the latter three higher is better.
Our framework leads to the best results on all six measures.
}
\label{fig:normal_table}
\end{table}
\begin{figure}[htb]
		\centering
		\setlength{\tabcolsep}{.15em}
        \renewcommand{\arraystretch}{1}
        \small
        \begin{tabular}{ccccccc} 
        &
        \rotatebox[origin=l]{90}{\centering CRN}&
        \includegraphics[width=.18\textwidth]{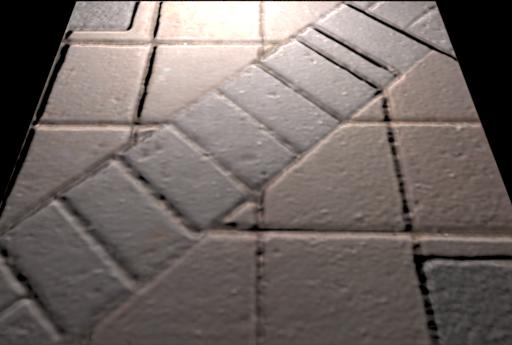}&
        \includegraphics[width=.18\textwidth]{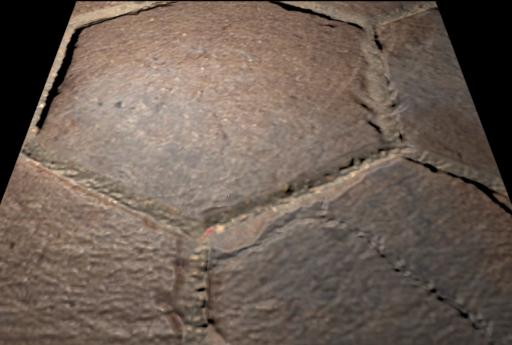}&
        \includegraphics[width=.18\textwidth]{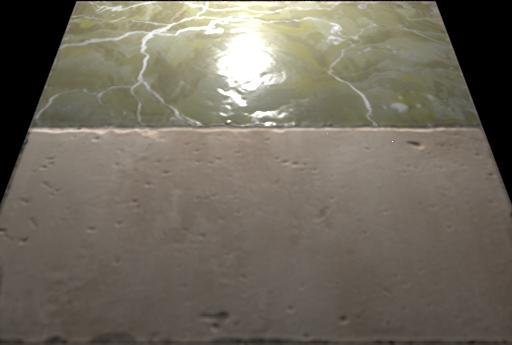}&       
        \includegraphics[width=.18\textwidth]{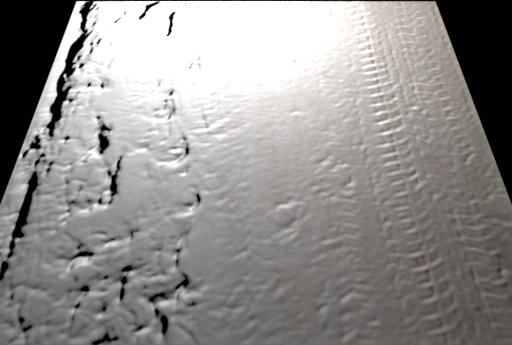}&
        \includegraphics[width=.18\textwidth]{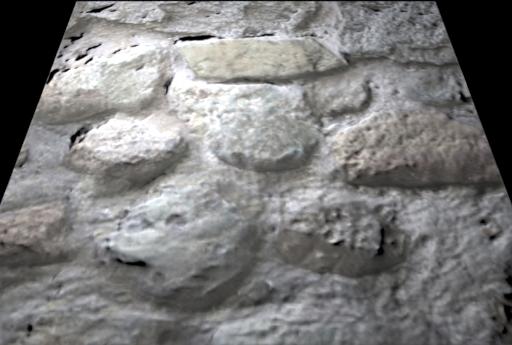}\\
        \rotatebox[origin=l]{90}{\centering \ \ \  Ours  }&
        \rotatebox[origin=l]{90}{\centering (with L1)}&
        \includegraphics[width=.18\textwidth]{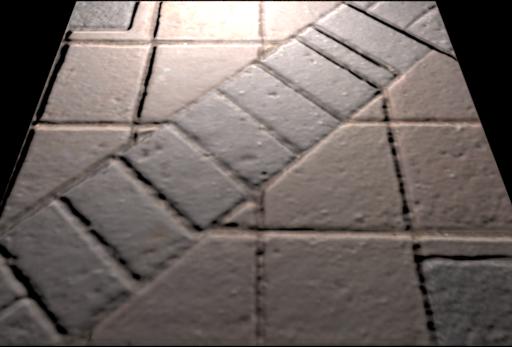}&
        \includegraphics[width=.18\textwidth]{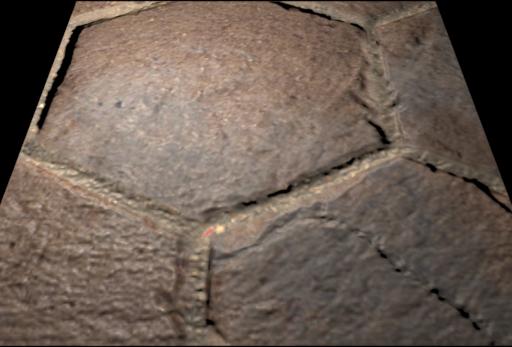}&
        \includegraphics[width=.18\textwidth]{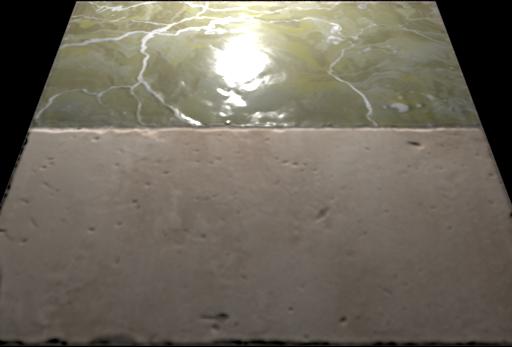}&       
        \includegraphics[width=.18\textwidth]{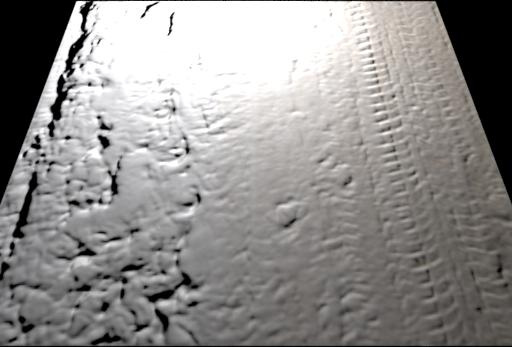}&
        \includegraphics[width=.18\textwidth]{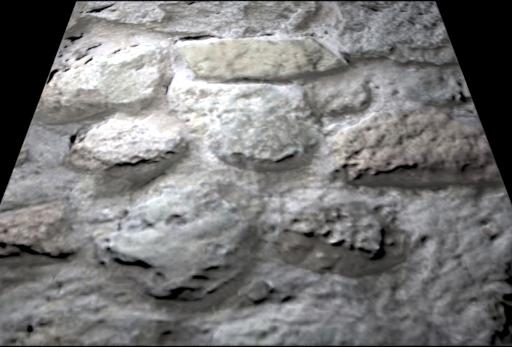}\\
        \rotatebox[origin=l]{90}{\centering \ \ \  Ours  }&
        \rotatebox[origin=l]{90}{\centering (w/o $L1$) }&
		\includegraphics[width=.18\textwidth]{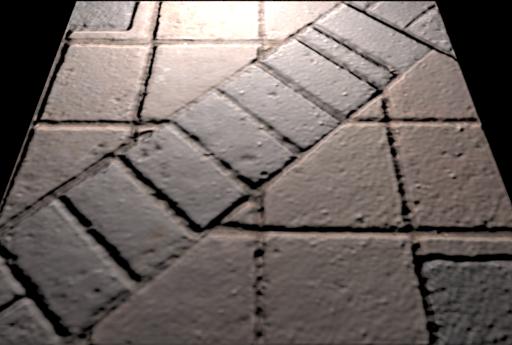}&
        \includegraphics[width=.18\textwidth]{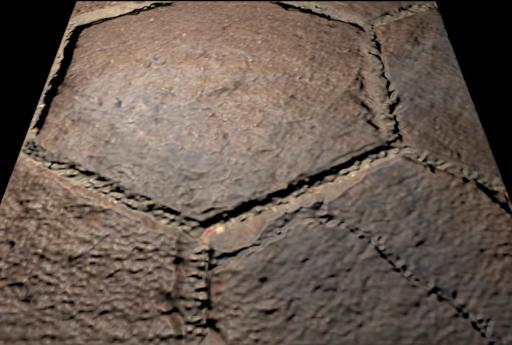}&
        \includegraphics[width=.18\textwidth]{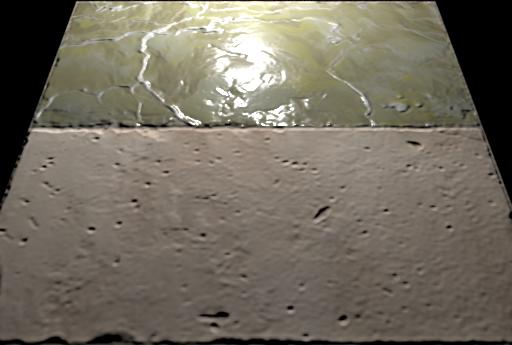}&
        \includegraphics[width=.18\textwidth]{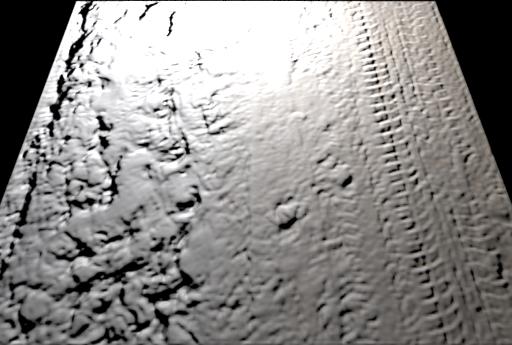}&
        \includegraphics[width=.18\textwidth]{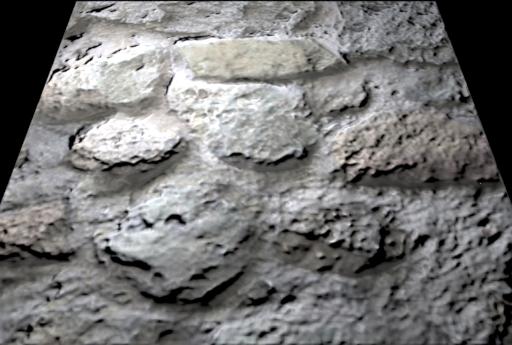}\\
        &
        \rotatebox[origin=l]{90}{\centering \ \ \ GT}&
		\includegraphics[width=.18\textwidth]{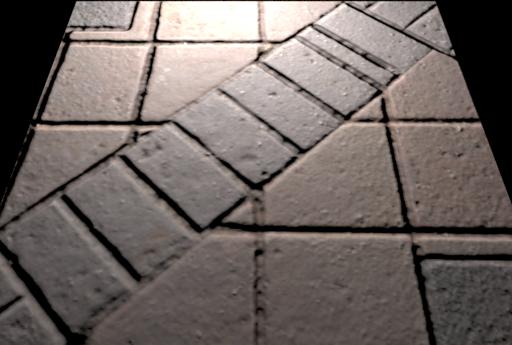}&
        \includegraphics[width=.18\textwidth]{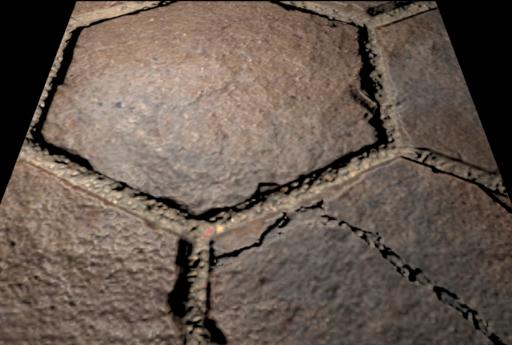}&
        \includegraphics[width=.18\textwidth]{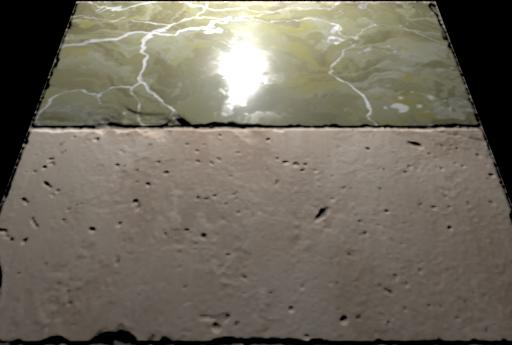}&
        \includegraphics[width=.18\textwidth]{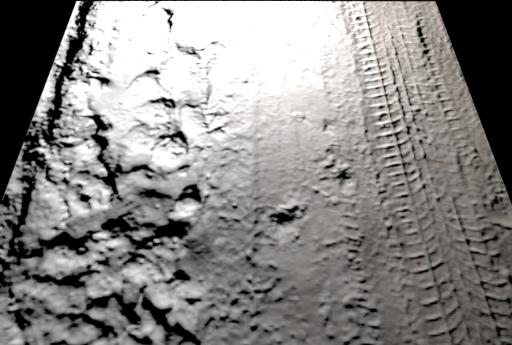}&
        \includegraphics[width=.18\textwidth]{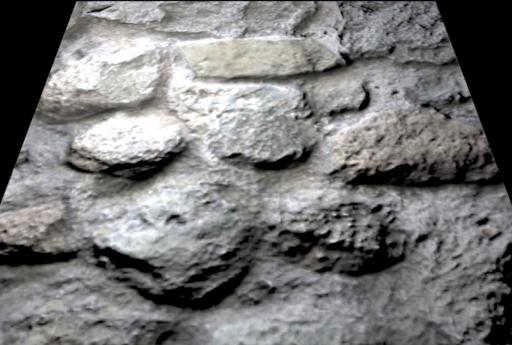}\\
        \end{tabular}
		\caption{\textbf{Rendered images}: To illustrate the benefits of our approach we render new view images using the estimated (or ground-truth) normal maps. Using our framework better maintains the statistics of the original surface and thus results with more natural looking images that are more similar to the ground-truth. Again, as was shown in Figure~\ref{fig:normal_data}, the results without $L1$ look more natural than those with $L1$.}
		\label{fig:rendered_data}
\end{figure}

\vspace*{-0.3cm}
We would like to draw your attention to the inferior scores obtained when removing the $L1$ term from our objective (i.e., setting $\lambda_{L1}\!=\!0$ in Eq.~\eqref{eq:normal-loss}).
Interestingly, this contradicts what one sees when examining the results visually. 
Looking at the reconstructed surfaces reveals that they look more natural and more similar to the ground-truth without $L1$.
A few examples illustrating this are provided in Figures~\ref{fig:normal_data},~\ref{fig:rendered_data}. 
This phenomena is actually not surprising at all. In fact, it is aligned with the recent trends in super-resolution, discussed in Section~\ref{sec:super-resolution}, where perceptual evaluation measures are becoming a common assessment tool. 
Unfortunately, such perceptual measures are not the common evaluation criteria for assessing reconstruction of surface normals.

Finally, we would like to emphasize, that our approach generalizes well to other textures outside our dataset. This can be seen from the results in Figure~\ref{fig:teaser} where two texture images, found online, were fed to our trained network. The reconstructed surface normals are highly detailed and look natural.

\section{Conclusions}
\label{sec:conclusion}
In this paper we proposed using loss functions based on a statistical comparison between the output and target for training generator networks.
It was shown via multiple experiments that such an approach can produce high-quality and state-of-the-art results.
While we suggest adopting the Contextual loss to measure the difference between distributions, other loss functions of a similar nature could (and should, may we add) be explored.
We plan to delve into this in our future work.

\noindent \textbf{Acknowledgements\ \ \ }
the Israel Science Foundation under Grant 1089/16 and
by the Ollendorf foundation.


\bibliographystyle{splncs}
\bibliography{egbib}

\begin{thebibliography}{10}

\bibitem{ruderman1994statistics}
Ruderman, D.L.:
\newblock The statistics of natural images.
\newblock Network: computation in neural systems (1994)

\bibitem{levin2003learning}
Levin, A., Zomet, A., Weiss, Y.:
\newblock Learning how to inpaint from global image statistics.
\newblock In: ICCV. (2003)

\bibitem{weiss2007makes}
Weiss, Y., Freeman, W.T.:
\newblock What makes a good model of natural images?
\newblock In: CVPR. (2007)

\bibitem{roth2009fields}
Roth, S., Black, M.J.:
\newblock Fields of experts.
\newblock IJCV (2009)

\bibitem{zoran2011learning}
Zoran, D., Weiss, Y.:
\newblock From learning models of natural image patches to whole image
  restoration.
\newblock In: ICCV, IEEE (2011)

\bibitem{goodfellow2014generative}
Goodfellow, I., Pouget-Abadie, J., Mirza, M., Xu, B., Warde-Farley, D., Ozair,
  S., Courville, A., Bengio, Y.:
\newblock Generative adversarial nets.
\newblock In: NIPS. (2014)

\bibitem{radford2015unsupervised}
Radford, A., Metz, L., Chintala, S.:
\newblock Unsupervised representation learning with deep convolutional
  generative adversarial networks.
\newblock In: ICLR. (2015)

\bibitem{arjovsky2017wasserstein}
Arjovsky, M., Chintala, S., Bottou, L.:
\newblock Wasserstein gan.
\newblock arXiv preprint arXiv:1701.07875 (2017)

\bibitem{ledig2016photo}
Ledig, C., Theis, L., Husz{\'a}r, F., Caballero, J., Cunningham, A., Acosta,
  A., Aitken, A., Tejani, A., Totz, J., Wang, Z.,  et~al.:
\newblock Photo-realistic single image super-resolution using a generative
  adversarial network.
\newblock In: CVPR. (2017)

\bibitem{isola2016image}
Isola, P., Zhu, J.Y., Zhou, T., Efros, A.A.:
\newblock Image-to-image translation with conditional adversarial networks.
\newblock In: CVPR. (2017)

\bibitem{salimans2016improved}
Salimans, T., Goodfellow, I., Zaremba, W., Cheung, V., Radford, A., Chen, X.:
\newblock Improved techniques for training gans.
\newblock In: NIPS. (2016)

\bibitem{levin2007blind}
Levin, A.:
\newblock Blind motion deblurring using image statistics.
\newblock In: Advances in Neural Information Processing Systems. (2007)
  841--848

\bibitem{aharon2006rm}
Aharon, M., Elad, M., Bruckstein, A.:
\newblock K-svd: An algorithm for designing overcomplete dictionaries for
  sparse representation.
\newblock IEEE Transactions on signal processing (2006)

\bibitem{mairal2010online}
Mairal, J., Bach, F., Ponce, J., Sapiro, G.:
\newblock Online learning for matrix factorization and sparse coding.
\newblock Journal of Machine Learning Research (2010)

\bibitem{hassner2006example}
Hassner, T., Basri, R.:
\newblock Example based 3d reconstruction from single 2d images.
\newblock In: CVPR Workshop. (2006)

\bibitem{michaeli2014blind}
Michaeli, T., Irani, M.:
\newblock Blind deblurring using internal patch recurrence.
\newblock In: ECCV. (2014)

\bibitem{glasner2009super}
Glasner, D., Bagon, S., Irani, M.:
\newblock Super-resolution from a single image.
\newblock In: ICCV. (2009)

\bibitem{zontak2011internal}
Zontak, M., Irani, M.:
\newblock Internal statistics of a single natural image.
\newblock In: CVPR. (2011)

\bibitem{mechrez2018contextual}
Mechrez, R., Talmi, I., Zelnik-Manor, L.:
\newblock The contextual loss for image transformation with non-aligned data.
\newblock (2018)

\bibitem{gal2007surface}
Gal, R., Shamir, A., Hassner, T., Pauly, M., Cohen-Or, D.:
\newblock Surface reconstruction using local shape priors.
\newblock In: Symposium on Geometry Processing. (2007)

\bibitem{huang2000statistics}
Huang, J., Lee, A.B., Mumford, D.:
\newblock Statistics of range images.
\newblock In: CVPR. (2000)

\bibitem{barrow1977parametric}
Barrow, H.G., Tenenbaum, J.M., Bolles, R.C., Wolf, H.C.:
\newblock Parametric correspondence and chamfer matching: Two new techniques
  for image matching.
\newblock Technical report (1977)

\bibitem{sun2018pix3d}
Sun, X., Wu, J., Zhang, X., Zhang, Z., Zhang, C., Xue, T., Tenenbaum, J.B.,
  Freeman, W.T.:
\newblock Pix3d: Dataset and methods for single-image 3d shape modeling.
\newblock In: CVPR. (2018)

\bibitem{de2012vision}
de~Villiers, H.A., van Zijl, L., Niesler, T.R.:
\newblock Vision-based hand pose estimation through similarity search using the
  earth mover's distance.
\newblock IET computer vision (2012)

\bibitem{botev2010kernel}
Botev, Z.I., Grotowski, J.F., Kroese, D.P.,  et~al.:
\newblock Kernel density estimation via diffusion.
\newblock The annals of Statistics (2010)

\bibitem{zhang2018unreasonable}
Zhang, R., Isola, P., Efros, A.A., Shechtman, E., Wang, O.:
\newblock The unreasonable effectiveness of deep features as a perceptual
  metric.
\newblock (2018)

\bibitem{wang2004image}
Wang, Z., Bovik, A.C., Sheikh, H.R., Simoncelli, E.P.:
\newblock Image quality assessment: from error visibility to structural
  similarity.
\newblock IEEE transactions on image processing (2004)

\bibitem{timofte2017NTIRE}
Timofte, R., Agustsson, E., Van~Gool, L., Yang, M.H., Zhang, L.,  et~al.:
\newblock Ntire 2017 challenge on single image super-resolution: Methods and
  results.
\newblock In: CVPR Workshops. (2017)

\bibitem{Lim2017edsr}
Lim, B., Son, S., Kim, H., Nah, S., Lee, K.M.:
\newblock Enhanced deep residual networks for single image super-resolution.
\newblock In: CVPR Workshops. (2017)

\bibitem{tong2017image}
Tong, T., Li, G., Liu, X., Gao, Q.:
\newblock Image super-resolution using dense skip connections.
\newblock In: CVPR. (2017)

\bibitem{blau2017perception}
Blau, Y., Michaeli, T.:
\newblock The perception-distortion tradeoff.
\newblock In: CVPR. (2018)

\bibitem{johnson2016perceptual}
Johnson, J., Alahi, A., Fei-Fei, L.:
\newblock Perceptual losses for real-time style transfer and super-resolution.
\newblock In: ECCV. (2016)

\bibitem{gatys2016image}
Gatys, L.A., Ecker, A.S., Bethge, M.:
\newblock Image style transfer using convolutional neural networks.
\newblock In: CVPR. (2016)

\bibitem{sajjadi2017enhancenet}
Sajjadi, M.S., Scholkopf, B., Hirsch, M.:
\newblock Enhancenet: Single image super-resolution through automated texture
  synthesis.
\newblock In: ICCV. (2017)

\bibitem{simonyan2014very}
Simonyan, K., Zisserman, A.:
\newblock Very deep convolutional networks for large-scale image recognition.
\newblock arXiv preprint arXiv:1409.1556 (2014)

\bibitem{agustsson2017div2k}
Agustsson, E., Timofte, R.:
\newblock Ntire 2017 challenge on single image super-resolution: Dataset and
  study.
\newblock In: CVPR Workshops. (2017)

\bibitem{martin2001database}
Martin, D., Fowlkes, C., Tal, D., Malik, J.:
\newblock A database of human segmented natural images and its application to
  evaluating segmentation algorithms and measuring ecological statistics.
\newblock In: ICCV, IEEE (2001)

\bibitem{ma2017learning}
Ma, C., Yang, C.Y., Yang, X., Yang, M.H.:
\newblock Learning a no-reference quality metric for single-image
  super-resolution.
\newblock Computer Vision and Image Understanding (2017)

\bibitem{chen2017photographic}
Chen, Q., Koltun, V.:
\newblock Photographic image synthesis with cascaded refinement networks.
\newblock In: ICCV. (2017)

\bibitem{bansal2017pixelnet}
Bansal, A., Chen, X., Russell, B., Ramanan, A.G.,  et~al.:
\newblock Pixelnet: Representation of the pixels, by the pixels, and for the
  pixels.
\newblock In: CVPR. (2016)

\bibitem{wang2015designing}
Wang, X., Fouhey, D., Gupta, A.:
\newblock Designing deep networks for surface normal estimation.
\newblock In: CVPR. (2015)

\bibitem{eigen2015predicting}
Eigen, D., Fergus, R.:
\newblock Predicting depth, surface normals and semantic labels with a common
  multi-scale convolutional architecture.
\newblock In: Proceedings of the IEEE International Conference on Computer
  Vision. (2015)  2650--2658

\bibitem{chen2017surface}
Chen, W., Xiang, D., Deng, J.:
\newblock Surface normals in the wild.
\newblock In: ICCV. (2017)

\bibitem{silbermanECCV12NYU}
Nathan~Silberman, Derek~Hoiem, P.K., Fergus, R.:
\newblock Indoor segmentation and support inference from rgbd images.
\newblock In: ECCV. (2012)

\end{thebibliography}
\end{document}